# LC-SEPLM: long-range contact-supervised adaptation for sequence-only protein representation learning

Chen Wang[1#], Boming Kang[2#], Qinghua Cui[2,3*]

[1]School of Electrical and Electronic Engineering, Hubei University of Technology, Hubei, China

[2]Department of Biomedical Informatics, State Key Laboratory of Vascular Homeostasis and Remodeling, School of Basic Medical Sciences, Peking University, 38 Xueyuan Rd, Beijing, 100191, China.

[3]School of Sports Medicine, Wuhan Institute of Physical Education, No. 461 Luoyu Rd. Wuchang District, Wuhan 430079, Hubei Province, China

# These authors contributed equally to this work

* To whom the correspondence should be addressed:

Qinghua Cui, email: cuiqinghua@bjmu.edu.cn

## Abstract

Protein language models learn transferable sequence representations. However, because they primarily model contextual dependencies along amino-acid sequences, their training objectives do not explicitly constrain the model to learn three-dimensional residue contacts formed after folding . Here, we introduce LC-SEPLM (Long-range Contact-supervised ESM Protein Language Model), which adapts ESM2 with LoRA and long-range residue-pair contact supervision while retaining sequence-only downstream inference. Pair-specific queries use cross-attention over the complete sequence to extract global sequence context associated with long-range spatial contacts. To expose the model to diverse structural information, we trained LC-SEPLM on 500,000 AlphaFold Swiss-Prot proteins. In downstream evaluation, LC-SEPLM improved all eight protein-level tasks relative to ESM2. The largest gain occurred in remote-homology recognition, where macro-F1 increased from 0.6122 to 0.6769 (+0.0647, or 6.47 percentage points). On the official ESM-S EC benchmark, LC-SEPLM also outperformed ESM-S with a maximum absolute gain of 0.1771. These results support residue-pair contact supervision as a bounded route for introducing structural information into protein sequence representations while preserving sequence-only inference.

## Introduction

Protein language models (pLMs) have emerged as a general framework for learning transferable representations from amino-acid sequences. Built on the Transformer architecture and masked language modeling (MLM) [1,2], pLMs are trained to recover masked residues from their surrounding sequence context. Training increasingly large models on diverse protein sequence collections has produced representations that encode information associated with protein structure, function and evolutionary relationships [3,4]. These representations have supported protein engineering, functional annotation and mutation-effect prediction [5–10]. However, although Transformers can integrate non-local sequence context, MLM directly supervises residue identity rather than structure-derived relations between residue pairs.

This distinction is important because protein folding brings residues that are distant along the primary sequence into spatial proximity. These long-range contacts contribute to tertiary organization, domain packing and the formation of functional environments [11,12]. Sequence-based pretraining may recover some of these relationships indirectly because evolutionary constraints affect the distribution of amino acids across protein families. However, predicting masked residues does not explicitly require the encoder to identify which sequence-distant residue pairs become spatially adjacent after folding. A central challenge is therefore to introduce direct structural supervision without requiring structural coordinates or derived structure representations during downstream deployment.

Existing approaches incorporate protein structure at several levels of representation. Graph-based and geometric models encode residue contacts, orientations or coordinates through graph neural networks, geometric message passing or structure-aware attention [13–17]. Structure-aware language models instead represent local tertiary environments as discrete tokens. SaProt combines amino-acid identities with Foldseek 3Di states, whereas ProstT5 learns transformations between protein sequences and structure-derived token sequences [18–20]. These methods provide direct access to structural information but generally require a structure, a predicted structure or a structure-derived representation as part of the model input or processing pipeline. A complementary strategy is to transfer structural knowledge into a sequence encoder during adaptation. ESM-S follows this direction by using fold- and remote-homology-related objectives while retaining sequence-only downstream inference [21]. Structure-conditioned sequence design and inverse-folding models provide further routes for learning sequence–structure relationships, although their inference setting differs from sequence representation learning [22,23].

Together, these studies demonstrate the value of structural information but leave the appropriate granularity of structural supervision unresolved. A protein structure can be represented as a whole-protein fold label, a per-residue structural state, a residue graph or an explicit relation between two residues. Protein- or domain-level objectives provide global supervision but do not directly identify the residue pairs that establish spatial constraints. Per-residue structure tokens encode local tertiary environments but compress pairwise relationships into independent positional states. Long-range contact labels retain the identity of both residues and directly represent whether two sequence-distant positions become spatially connected. Their use is computationally challenging because a protein of length$(L)$ contains $(O(L^2))$ candidate residue pairs, requiring controlled pair sampling and an objective focused on informative long-range relations.

Here we introduce LC-SEPLM, the Long-range Contact-supervised ESM Protein Language Model. LC-SEPLM adapts ESM2 using low-rank adaptation (LoRA) [24] and residue-pair contact labels derived from AlphaFold structures of Swiss-Prot proteins [25–27]. For each sampled residue pair, the model combines the two residue states, their representation-level interactions and their sequence separation into a pair-specific query. This query attends to the complete sequence, allowing contact prediction to incorporate the global sequence context associated with the selected pair. LC-SEPLM does not claim parameter-free adaptation. LoRA introduces additional trainable parameters that remain in the adapted encoder, whereas the pair-query, cross-attention and contact-prediction modules form a training-only structural supervision branch. This branch is removed after adaptation, so downstream inference uses the adapted ESM2 encoder and amino-acid sequence alone, without coordinates, contact maps, residue graphs or structure-derived tokens.

We trained LC-SEPLM on 500,000 AlphaFold Swiss-Prot proteins and evaluated the frozen adapted encoder on eight protein-level prediction tasks and two residue- or site-level diagnostics. Under the reported downstream evaluation protocol, LC-SEPLM improved all eight protein-level tasks relative to ESM2, with the strongest gain observed for remote-homology recognition. The transfer benefit was not universal: neither ProteinGym mutation-effect prediction nor post-translational-modification site prediction improved. Evaluation on the independent ESM-S benchmark further showed consistent gains for Enzyme Commission prediction but mixed results for fold classification and Gene Ontology molecular function. These findings position long-range residue-pair supervision as a practical but bounded strategy for enriching sequence representations with structural information while preserving sequence-only downstream deployment.

## Results

### Evaluation design

Figure 1 summarizes the LC-SEPLM workflow. AlphaFold-derived long-range contact labels were used only during adaptation, whereas downstream evaluation used the adapted sequence encoder without structural coordinates, the contact head or the auxiliary MLM head. The controlled comparison comprised eight protein-level tasks and two residue- or site-level diagnostics (Table 1).

The protein-level datasets ranged from 9,817 proteins for EC level 1 prediction to 189,349 proteins for Pfam family classification and covered binary, multiclass and multilabel outputs. ProteinGym mutation-effect prediction and UniProt PTM-site prediction were analysed separately because they probe local variant- or residue-level behaviour rather than pooled protein embeddings.

### Contact supervision converged while retaining sequence modeling

The final Tail-8 + MLM 0.6 model was trained for 14 epochs. Training and validation total loss declined overall, validation contact balanced accuracy increased from 0.9421 at epoch 1 to 0.9630 at epoch 12, and validation MLM accuracy remained near 0.48 (Figure 2). Thus, the long-range contact objective was learned without collapse of the auxiliary sequence-modeling objective.

The best upstream contact score occurred at epoch 12, whereas downstream validation selected epoch 10. We therefore treated contact balanced accuracy as an adaptation diagnostic rather than as the sole criterion for choosing the transferable encoder.

**LC-SEPLM improved all eight protein-level tasks**

At the downstream-selected epoch-10 checkpoint, LC-SEPLM outperformed ESM2 on all eight protein-level tasks (Figure 3 and Table 2). The largest gain occurred in Pfam-clan remote-homology recognition, where macro-F1 increased from 0.6122 to 0.6769 (absolute difference, 0.0647). GO-MF, Pfam family and strict membrane classification improved by 0.0098, 0.0056 and 0.0042, respectively.

The remaining protein-level gains were smaller than 0.003: EC increased by 0.0015, subcellular localization by 0.0020, membrane classification by 0.0004 and HOU level 3 by 0.0022. By contrast, ProteinGym mean Spearman correlation decreased from 0.410952 to 0.403334, and PTM-site mean AUROC changed from 0.777625 to 0.777483. The transfer benefit was therefore broad at the protein level but did not extend to the two local diagnostics.

Figure 8 shows the same final Tail-8 comparison as absolute task scores with downstream variability. Each protein-level task used five classifier seeds except GO-MF, which used the three-seed validation gate. This detailed view confirms that the positive signs in Figure 3 are derived from the final epoch-10 encoder rather than from the early all-layer series.

Remote homology remained above the ESM2 baseline across all 14 evaluated random-split checkpoints (Figure 4). The epoch-10 mean was 0.6769, matching Table 2, whereas the highest random-split mean occurred at epoch 14. The persistence of the gain across checkpoints indicates that the remote-homology result was not confined to a single model snapshot.

**Independent ESM-S comparison showed the strongest gains for EC prediction**

On the official ESM-S benchmark, LC-SEPLM exceeded ESM-S 150M on all three EC identity splits, with F1-max gains of 0.1389, 0.1506 and 0.1771 (Table 3). Fold-family macro-F1 also increased by 0.0121.

The external comparison was not uniformly positive. Fold-superfamily and fold-level scores decreased by 0.0286 and 0.0078, respectively, and GO-MF decreased on all three identity splits. These results support an EC-specific advantage but do not establish general superiority over ESM-S.

**Downstream validation, rather than contact accuracy, selected the checkpoint**

Checkpoint analysis used only the final Tail-8 series (Figure 5). Validation contact balanced accuracy was 0.9613, 0.9630 and 0.9626 at epochs 10, 12 and 14, respectively. The downstream gate evaluated EC and GO-MF over three gate seeds and ProteinGym over 93 development assays.

At epoch 10, the EC, GO-MF and ProteinGym changes relative to ESM2 were +0.0191, +0.0132 and −0.0076. The corresponding values were +0.0161, +0.0099 and −0.0091 at epoch 12, and +0.0172, +0.0052 and −0.0085 at epoch 14. Among these prespecified late checkpoints, epoch 10 gave the strongest gate profile even though the upstream contact metric peaked at epoch 12.

The gate and the final comparison serve different purposes. Figure 5 reports validation-gate values used for checkpoint selection, whereas Table 2 and Figures 3 and 8 report the final task-specific comparison; values from these evaluation stages should not be interpreted as estimates from the same split.

**Supporting configuration and filtering analyses**

Accession and sequence-identity filtering was audited for seven curated protein-level tasks (Figure 6). Exact-accession filtering retained 6,949 task-level evaluation entries (9.0% of the pooled original sets), whereas 40% and 30% identity filtering retained 1,607 (2.1%) and 1,313 (1.7%) entries. These are summed task-level entries rather than unique proteins. HOU was not included in this filtering package because it has no independent validation partition.

The LoRA-depth sweep fixed the MLM weight at 0.3 and trained each candidate for one epoch; the MLM-weight sweep fixed Tail-8 and trained each candidate for two epochs (Figure 7). Both sweeps reported the mean change across eight protein-level tasks, PTM mean AUPRC and ProteinGym mean Spearman correlation. Tail-8 and an MLM weight of 0.6 were selected as the balanced operating point before the 14-epoch final run; these model-selection sweeps are not independent confirmatory tests.

Table 4 records the early all-30 and final Tail-8 LoRA configurations, and Table 5 lists the curated splits for all eight protein-level tasks, including HOU with no validation partition. Table 6 positions LC-SEPLM relative to other structure-aware strategies. Table 7 retains the excluded ClinVar and metal-ion outputs as diagnostic audits without including them in the primary performance claim.

## Discussion

LC-SEPLM shows that long-range residue-pair contact supervision can improve a sequence-only pLM encoder across a broad set of protein-level tasks. The largest effect appeared in remote-homology recognition, a task that benefits from structural relationships that may be weakly reflected in sequence similarity.

The results also define the boundary of the approach. LC-SEPLM did not improve ProteinGym variant-effect prediction or PTM-site prediction, and it underperformed ESM-S on several non-EC benchmark splits. These outcomes argue against a universal improvement claim.

A plausible interpretation is that long-range contact supervision strengthens global protein representations more than local residue-level signals. This interpretation is consistent with the task pattern observed here, but it requires additional testing on datasets designed to separate global fold information from local biochemical effects.

The use of predicted AlphaFold structures is both a strength and a limitation. It enables large-scale supervision, but it may transfer prediction biases from the structure source into LC-SEPLM. Experimental structures or confidence-aware contact weighting could test the robustness of this supervision.

The comparison with ESM-S should also be interpreted carefully. LC-SEPLM was evaluated against official ESM-S scores rather than retrained under an identical pipeline, so differences may reflect both representation quality and benchmark design.

Future work should test larger base encoders, alternative contact definitions and multi-objective schedules. It should also evaluate whether contact supervision can be combined with residue-level objectives to improve local prediction tasks.

Overall, the results support LC-SEPLM as a bounded adaptation strategy for sequence-only protein representation learning. The method is most compelling when the downstream task depends on global structural context and least compelling when the task is dominated by local residue effects.

In conclusion, LC-SEPLM introduces long-range residue-pair contact supervision into ESM2 adaptation while retaining sequence-only downstream inference. Across the controlled ESM2 comparison, the method improved all eight protein-level tasks and achieved its largest gain in remote-homology recognition. The absence of improvement on ProteinGym and PTM-site diagnostics shows that the benefit is not universal. These results support contact-supervised adaptation as a practical way to enrich global protein representations, with clear limits for local prediction tasks.

## Materials and methods

### Overview of LC-SEPLM

LC-SEPLM adapts ESM2-t30-150M-UR50D with long-range contact supervision while preserving sequence-only inference. Given a protein sequence $S = (s_1, \dots, s_L)$, the adapted encoder $f_\theta(S)$ is trained to improve protein-level representations. During adaptation, structure-derived labels $y_{ij}$ are available for sampled residue pairs (i, j). The overall workflow is summarized in Figure 1.

$$y_{ij} = \mathbb{I}\left(|i - j| \geq 12 \ \wedge \ d_{ij}^{C\alpha} \leq 8\,\text{Å}\right)$$

Here, $d_{ij}^{C}$ denotes the Cα to Cα distance. The sequence-separation threshold removes local sequential contacts and focuses the auxiliary task on long-range spatial relations that are more informative for global fold organization.

### Contact supervision and LoRA adaptation

The backbone encoder was ESM2-t30-150M-UR50D with 30 Transformer layers [4]. LoRA modules were inserted into attention projection matrices, and the frozen ESM2 weights were retained. For a frozen matrix $W_0$, the adapted matrix was parameterized as:

$$W_{\text{adapted}} = W_0 + \Delta\text{W}$$

$$\Delta\text{W} = \frac{\alpha}{r} BA$$

$$y = W_0 x + \frac{\alpha}{r} BAx$$

A and B are trainable low-rank matrices, with $r = 4, \alpha = 8.0$ and LoRA dropout = 0.05. The final configuration used LoRA modules in the last eight Transformer layers; a configuration adapting all 30 layers was retained only as an early diagnostic comparison.

## Training data and contact-label construction

Upstream adaptation used AlphaFold-predicted structures for Swiss-Prot proteins [25–27]. The data manifest generated with seed 23 contained 535,750 training proteins, 5,629 validation proteins and 5,589 test proteins. The final run sampled 500,000 training proteins and 2,000 proteins from each held-out split, following the recorded run metadata.

Structural coordinates were used only to construct training labels. Sequences were truncated to 512 residues. A positive contact required a sequence separation of at least 12 residues and a Cα to Cα distance of at most 8 Å; all other sampled pairs were treated as negatives. At most 768 candidate pairs were sampled per protein with an approximate 2:1 negative-to-positive ratio, controlling the $O(L^2)$ pair space while retaining long-range positives.

## Pair-query contact head

Let the ESM2 encoder output residue states:

$$H = [h_1, h_2, \dots, h_L]$$

For each sampled pair (i, j), we concatenated the two residue states, their absolute difference, their element-wise product and normalized sequence distance:

$$x_{ij} = [h_i; h_j; |h_i - h_j|; h_i \odot h_j; \delta_{ij}]$$

$$\delta_{ij} = \frac{|i-j|}{L}$$

The pair descriptor has $4d + 1 = 2{,}561$ features for ESM2-150M, where $d = 640$. A pair-query MLP projected this descriptor to the ESM2 hidden width before cross-attention over the full residue sequence:

$$q_{ij} = \mathrm{MLP}_{\mathrm{pair}}(x_{ij})$$

$$Q_{ij} = q_{ij}W^Q, K = HW^K, V = HW^V$$

$$z_{ij} = \text{softmax}(\frac{Q_{ij}K^T}{\sqrt{d_k}})V$$

$$\hat{\text{p}}_{ij} = \text{sigmoid}(\text{MLP}_{\text{contact}}(z_{ij}))$$

This design asks a pair-specific query whether two sequence positions are spatially connected, rather than requiring structural coordinates during downstream inference.

### Optimization objective and checkpoint selection

LC-SEPLM was optimized with a binary contact objective and an auxiliary masked-language-modeling objective. For sampled residue-pair set P and masked-token set M, the losses were:

$$L_{\text{contact}} = -\frac{1}{|P|}\Sigma(i,j) \in P[y_{ij}\log(\hat{\text{p}}_{ij}) + (1 - y_{ij})\log(1 - \hat{\text{p}}_{ij})]$$

$$L_{\text{MLM}} = -\frac{1}{|M|}\Sigma k \in M \log p_\theta(s_k | S_{\text{masked}})$$

$$L_{\text{total}} = L_{\text{contact}} + \lambda_{\text{MLM}} L_{\text{MLM}}$$

The selected configuration used $\lambda_{\text{MLM}} = 0.6$ over a 14-epoch adaptation run. Checkpoint selection used downstream validation rather than contact accuracy alone, because contact prediction is an auxiliary adaptation signal and the intended output is a frozen sequence encoder.

### Downstream representation extraction

After adaptation, structural coordinates and the contact head were removed. Each downstream protein was encoded from sequence alone. We mean-pooled valid amino-acid token states while excluding padding, BOS and EOS tokens:

$$e_{protein} = \frac{1}{L}\sum_{i=1}^{L} h_i$$

The resulting 640-dimensional vector was used as the frozen protein embedding for downstream classifiers or regressors.

**Downstream evaluation tasks and metrics**

The controlled ESM2 comparison used eight protein-level tasks and two local diagnostics derived from established protein benchmark and database resources [9,10,27–32]. Protein-level tasks included EC prediction, subcellular localization, strict membrane classification, GO-MF prediction, Pfam family classification, remote-homology recognition and HOU localization. ProteinGym and PTM-site prediction were retained as local diagnostics because they probe variant- or residue-level behavior rather than global protein embeddings.

We also compared LC-SEPLM with ESM-S on the official ESM-S benchmark. Because the ESM-S task definitions and metrics differ from the controlled ESM2 benchmark, those results were analyzed separately. Classification tasks were reported with the metric defined by the corresponding benchmark, including macro-F1 or F1-max where applicable.

## Key Points

LC-SEPLM introduces long-range residue-pair contact supervision during ESM2 adaptation while retaining sequence-only downstream inference.

The adapted encoder improved all eight protein-level tasks relative to ESM2, with the largest gain in remote-homology recognition (+0.0647 macro-F1).

ProteinGym and post-translational modification-site diagnostics did not improve, indicating that the benefits were concentrated in global rather than local representations.

Contact-objective accuracy and downstream transfer peaked at different checkpoints, supporting validation across representative downstream tasks.

**Author contributions**

C.W. and B.K. conceived the study and designed the LC-SEPLM framework.

B.K. curated datasets, analyzed results, and prepared figures and tables.

C.W. implemented the model, performed experiments, contributed to data analysis, result interpretation, and manuscript drafting.

Q.C. supervised the study, provided conceptual guidance, and revised the manuscript.

All authors read and approved the final manuscript.

## Funding

**Conflict of interest**

The authors declare no competing interests.

**Data availability**

The data analysed in this study were obtained from publicly available protein sequence, structure and benchmark resources, including UniProt/Swiss-Prot, the AlphaFold Protein Structure Database and the downstream benchmark datasets described in the Materials and methods section and Tables 1 and 5. Processed data required to reproduce the main analyses are available from the corresponding author upon reasonable request.

**Code availability**

The source code and model-related files supporting LC-SEPLM will be made publicly available after publication.

# Tables

Table 1 | Downstream benchmark composition. The table lists the task type, prediction level, output space, dataset size, train-validation-test split and primary metric for the controlled ESM2 comparison. Protein-level tasks were analyzed separately from ProteinGym mutation-effect prediction and UniProt PTM-site prediction, which serve as local diagnostic readouts.

| Task | Prediction level | Output space | Total size | Train | Validation | Test | Primary metric |
|---|---|---|---|---|---|---|---|
| EC level 1 | Protein | 7 classes | 9,817 proteins | 7,859 | 977 | 981 | Macro-F1 |
| Subcellular localization | Protein | 11 classes | 12,374 proteins | 9,903 | 1,233 | 1,238 | Macro-F1 |
| Membrane classification | Protein | 2 classes | 20,000 proteins | 16,005 | 1,999 | 1,996 | Macro-F1 |
| Strict membrane classification | Protein | 2 classes | 20,000 proteins | 16,008 | 1,995 | 1,997 | Macro-F1 |
| GO molecular function | Protein | 40 labels | 30,000 proteins | 24,013 | 2,993 | 2,994 | Micro-F1 |
| Pfam family classification | Protein | 510 families | 189,349 proteins | 151,362 | 18,683 | 19,304 | Macro-F1 |
| Pfam-clan remote homology | Protein | 108 clans | 82,085 proteins | 61,563 | 8,005 | 12,517 | Macro-F1 |
| HOU subcellular localization level 3 | Protein | 7 labels | 20,887 proteins | 17,661 | — | 3,226 | Micro-F1 |
| ProteinGym mutation-effect prediction | Mutation/assay | 93 assays | 93 assays | — | Development panel | — | Mean Spearman correlation |
| UniProt PTM-site prediction | Residue/site | 3 PTM types | 250 proteins per subtype | 200 per subtype | — | 50 per subtype | Mean AUROC |

Table 2 | Controlled comparison between ESM2 and LC-SEPLM at the downstream-selected epoch-10 checkpoint. Values are mean ± s.d.; Delta is LC-SEPLM minus ESM2. GO-MF reports the three-seed validation gate, ProteinGym reports the 93-assay development panel, and the remaining replicate units are listed in the n column.

| Task | Metric | ESM2 | LC-SEPLM (epoch 10) | Delta | n |
|---|---|---|---|---|---|
| EC level 1 | macro-F1 | 0.883669 ± 0.002586 | 0.885186 ± 0.005664 | +0.001517 | 5 |
| Subcellular location | macro-F1 | 0.759775 ± 0.005115 | 0.761797 ± 0.002369 | +0.002022 | 5 |
| Membrane binary | macro-F1 | 0.917088 ± 0.004614 | 0.917534 ± 0.002996 | +0.000446 | 5 |
| Strict membrane binary | macro-F1 | 0.976577 ± 0.001426 | 0.980743 ± 0.000328 | +0.004166 | 5 |
| GO molecular function | micro-F1 | 0.709639 ± 0.005532 | 0.719465 ± 0.003628 | +0.009826 | 3 |
| Pfam family | macro-F1 | 0.978692 ± 0.000533 | 0.984260 ± 0.000363 | +0.005568 | 5 |
| Remote homology (Pfam clan) | macro-F1 | 0.612240 ± 0.003315 | 0.676943 ± 0.003424 | +0.064703 | 5 |
| HOU subcellular level 3 | micro-F1 | 0.531879 ± 0.018151 | 0.534068 ± 0.016696 | +0.002189 | 5 |
| ProteinGym mutation effect | mean Spearman | 0.410952 ± 0.192679 | 0.403334 ± 0.189175 | −0.007618 | 93 assays |
| UniProt PTM-site probes | mean AUROC | 0.777625 ± 0.009449 | 0.777483 ± 0.015431 | −0.000142 | 5 split seeds |

Table 3 | Independent comparison with the official ESM-S benchmark. LC-SEPLM scores are compared with reported ESM-S 150M results on EC, fold-classification and GO-MF splits. Delta is LC-SEPLM minus ESM-S 150M using the metric defined by each task. Deltas were calculated from unrounded source values; subtraction of the displayed four-decimal scores can therefore differ by 0.0001.

| Official task | Test split | ESM-S 150M | LC-SEPLM (epoch 3) | Delta (LC − ESM-S) |
|---|---|---|---|---|
| EC | <30% identity | 0.2553 | 0.3942 | +0.1389 |
| EC | <50% identity | 0.2636 | 0.4142 | +0.1506 |
| EC | <95% identity | 0.2880 | 0.4652 | +0.1771 |
| Fold | family | 0.9242 | 0.9364 | +0.0121 |
| Fold | superfamily | 0.4905 | 0.4619 | −0.0286 |
| Fold | fold | 0.1241 | 0.1163 | −0.0078 |
| GO-MF | <30% identity | 0.4882 | 0.4671 | −0.0211 |
| GO-MF | <50% identity | 0.5208 | 0.5015 | −0.0194 |
| GO-MF | <95% identity | 0.5739 | 0.5610 | −0.0129 |

Table 4 | LoRA adaptation configurations. The table compares the early all-layer diagnostic configuration with the final last-8-layer configuration, including adapted layers, target modules, rank, scaling factor and dropout.

| Configuration | LoRA layers | Target modules | Rank | Alpha | Dropout |
|---|---|---|---|---|---|
| Early all-30 | 0–29 | q_proj, v_proj | 4 | 8.0 | 0.05 |
| Final last-8-layer | 22–29 | q_proj, v_proj | 4 | 8.0 | 0.05 |

Table 5 | Curated downstream splits for all eight protein-level tasks. Identical sequences were assigned to a single partition. Seven tasks have train-validation-test splits; HOU level 3 has a predefined training and test set but no independent validation partition.

| Task | Train / val / test |
|---|---|
| EC | 7,859 / 977 / 981 |
| Subcellular localization | 9,903 / 1,233 / 1,238 |
| Membrane | 16,005 / 1,999 / 1,996 |
| Strict membrane | 16,008 / 1,995 / 1,997 |
| GO-MF | 24,013 / 2,993 / 2,994 |
| Pfam family | 151,362 / 18,683 / 19,304 |
| Remote homology | 61,563 / 8,005 / 12,517 |
| HOU subcellular level 3 | 17,661 / — / 3,226 |

Table 6 | Methodological positioning of LC-SEPLM relative to related structure-aware representation strategies. The table contrasts structural attention, structure-aware vocabularies, structural-task distillation and pairwise contact supervision according to their structural signal and downstream structural-input requirement.

| Route | Representative method | Structural signal | Downstream structural input |
|---|---|---|---|
| Structure-aware attention | Structure-Aware Transformer, PST | Local graph features modify attention | Usually required |
| Structure-aware vocabulary | SaProt | Amino-acid token + 3Di token | Usually required for tokenization |
| Structural-task distillation | Structure-Informed PLM | Fold/remote-homology supervision | Not required |
| Pairwise contact supervision | LC-SEPLM | Long-range residue-pair contact labels | Not required |

Table 7 | Excluded diagnostic evaluations. ClinVar wild-type marginals and metal-ion binding probes were retained as diagnostic audits only because the archived outputs or near-zero scores did not support interpretable main-text performance comparisons.

| Diagnostic | ESM2 | ESM-S | LC-SEPLM | Decision and rationale |
|---|---|---|---|---|
| ClinVar wild-type marginals, mean Spearman (1,779 assays) | 0.0000 | 0.0000 | 0.0000 | Excluded; all archived assay scores are exactly zero for every model, so the evaluator/data path requires audit. |
| Metal-ion binding, accuracy | 0.007463 ± 0.000000 | 0.006866 ± 0.001335 | 0.007463 ± 0.000000 | Diagnostic only; near-zero output for all models is inconsistent with a valid binary probe. |
| Metal-ion binding, macro-F1 | 0.007526 ± 0.000369 | 0.006990 ± 0.001428 | 0.007587 ± 0.000444 | Diagnostic only; no model has interpretable discriminative performance. |

# Figure legends and figures

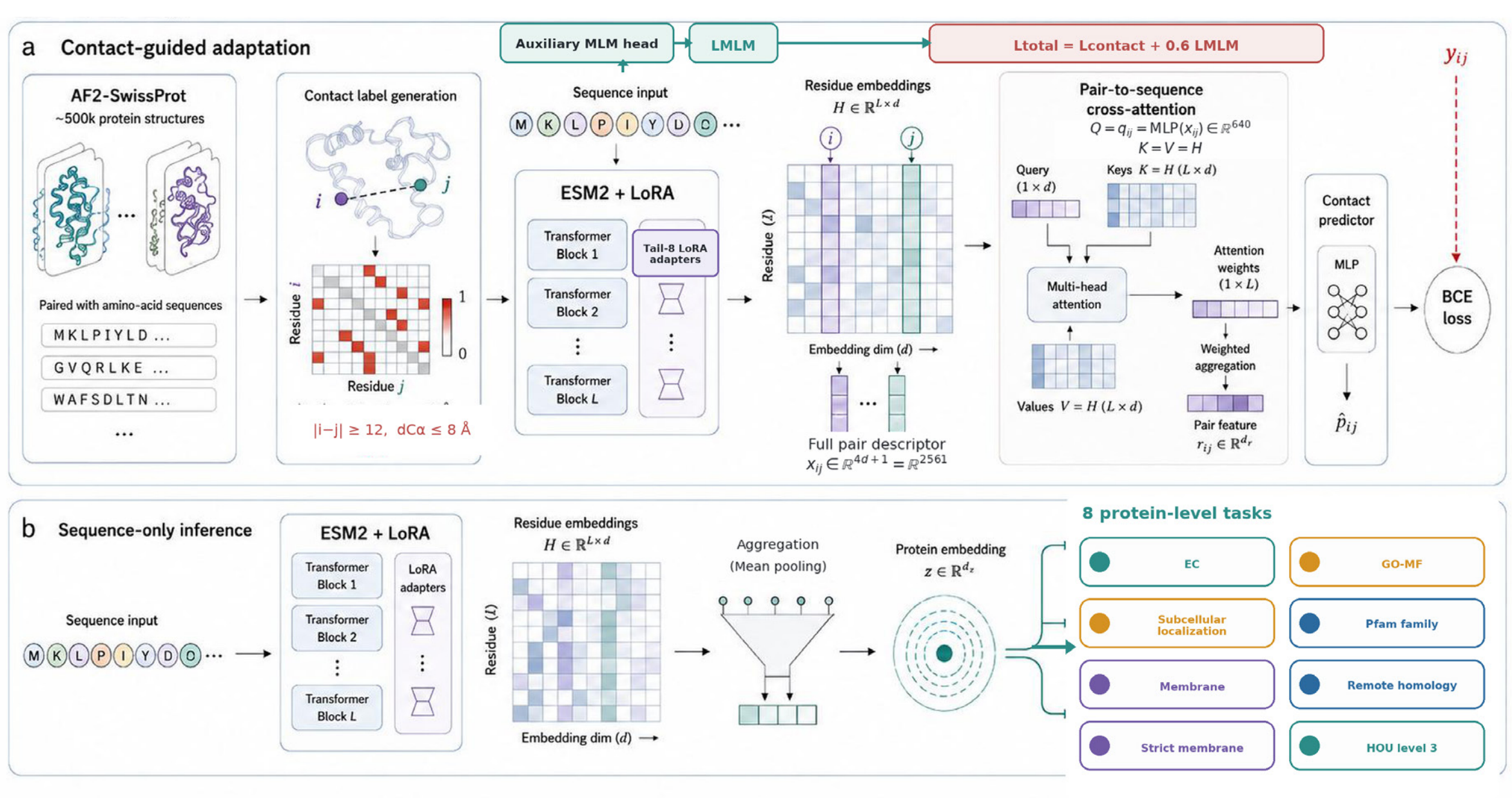

Figure 1 | LC-SEPLM adaptation and sequence-only inference. a, AlphaFold Swiss-Prot structures define long-range contacts by |i−j| ≥ 12 and Cα distance ≤ 8 Å. ESM2-150M is adapted through Tail-8 LoRA while a pair query attends to the full residue sequence. Contact loss is combined with an auxiliary MLM loss weighted by 0.6; the contact and MLM heads are used only during adaptation. b, Downstream inference uses amino-acid sequence alone. Mean pooling yields a 640-dimensional protein embedding for eight protein-level tasks: EC, GO-MF, subcellular localization, Pfam family, membrane classification, strict membrane classification, remote homology and HOU level 3.

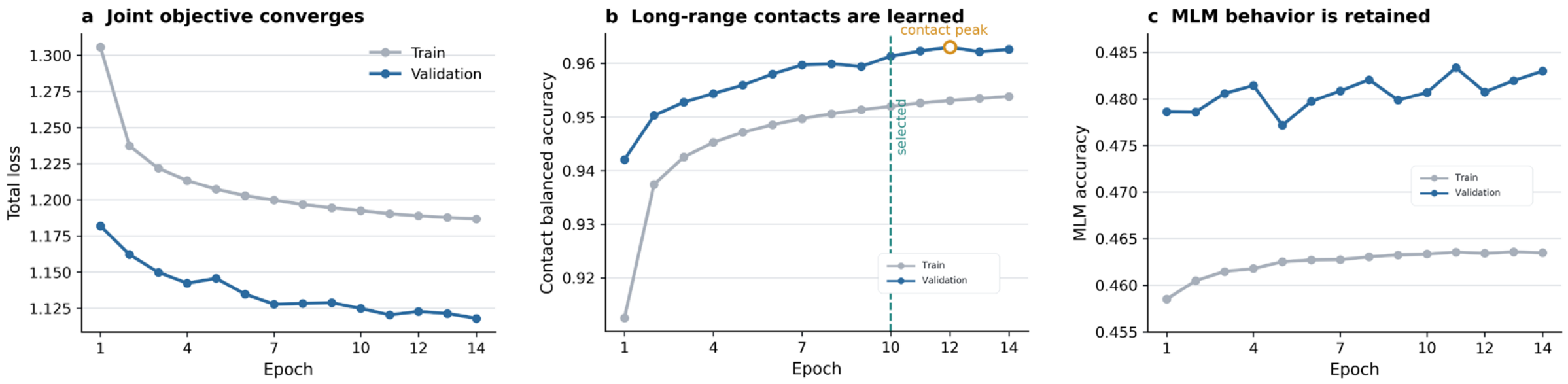

Figure 2 | Optimization of the final Tail-8 + MLM 0.6 model. a, Training and validation total loss across 14 epochs. b, Training and validation contact balanced accuracy; the vertical line marks the downstream-selected epoch 10 and the open circle marks the validation peak at epoch 12. c, Training and validation MLM accuracy. Gray denotes training and blue denotes validation. All panels use the single encoder-training run with seed 23.

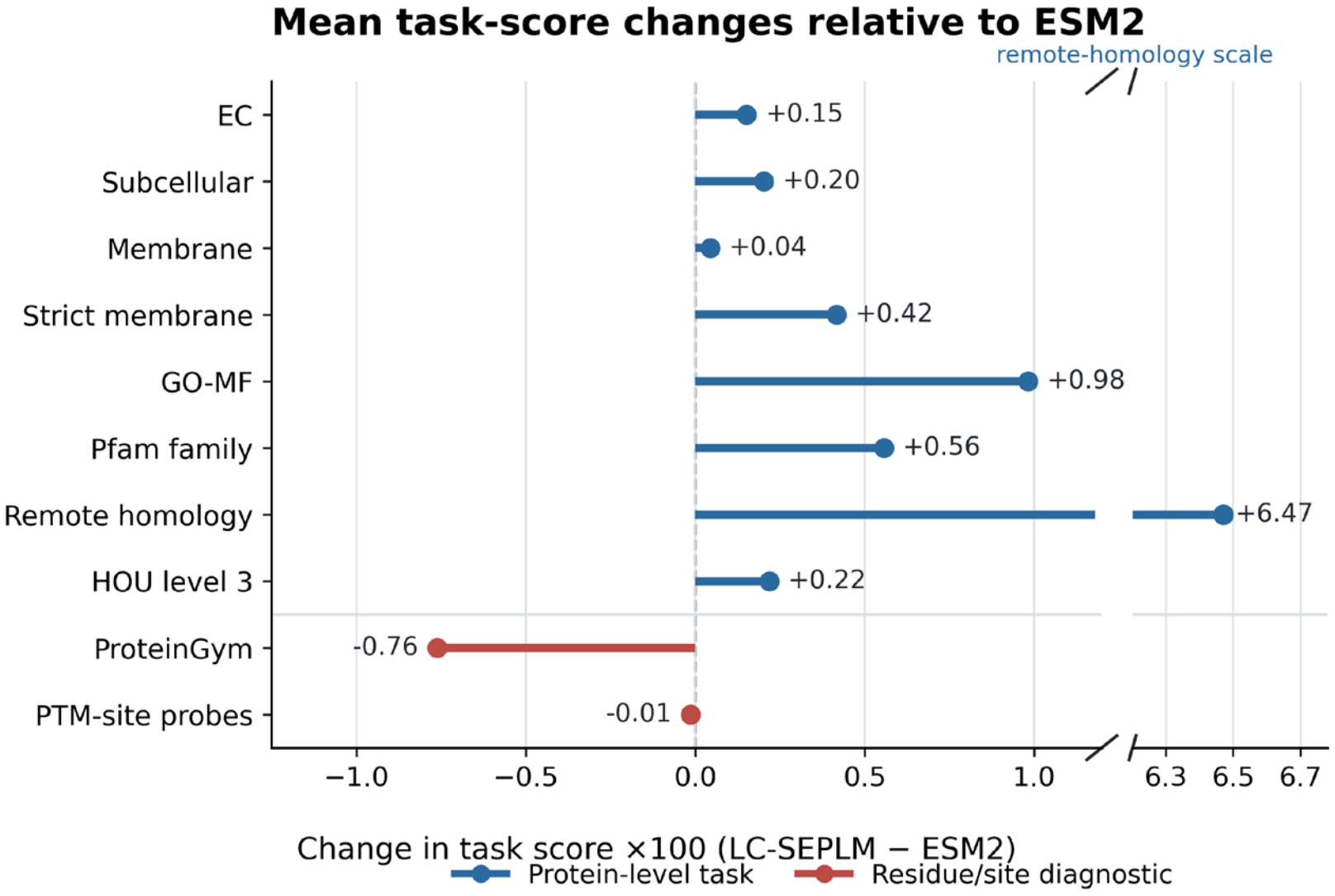

Figure 3 | Final epoch-10 task-score changes relative to ESM2. Blue denotes eight protein-level tasks and red denotes two residue- or site-level diagnostics. Values are changes in each task-specific score multiplied by 100 and correspond to Table 2. Remote homology uses a broken horizontal scale because its gain is substantially larger than the remaining task-level changes.

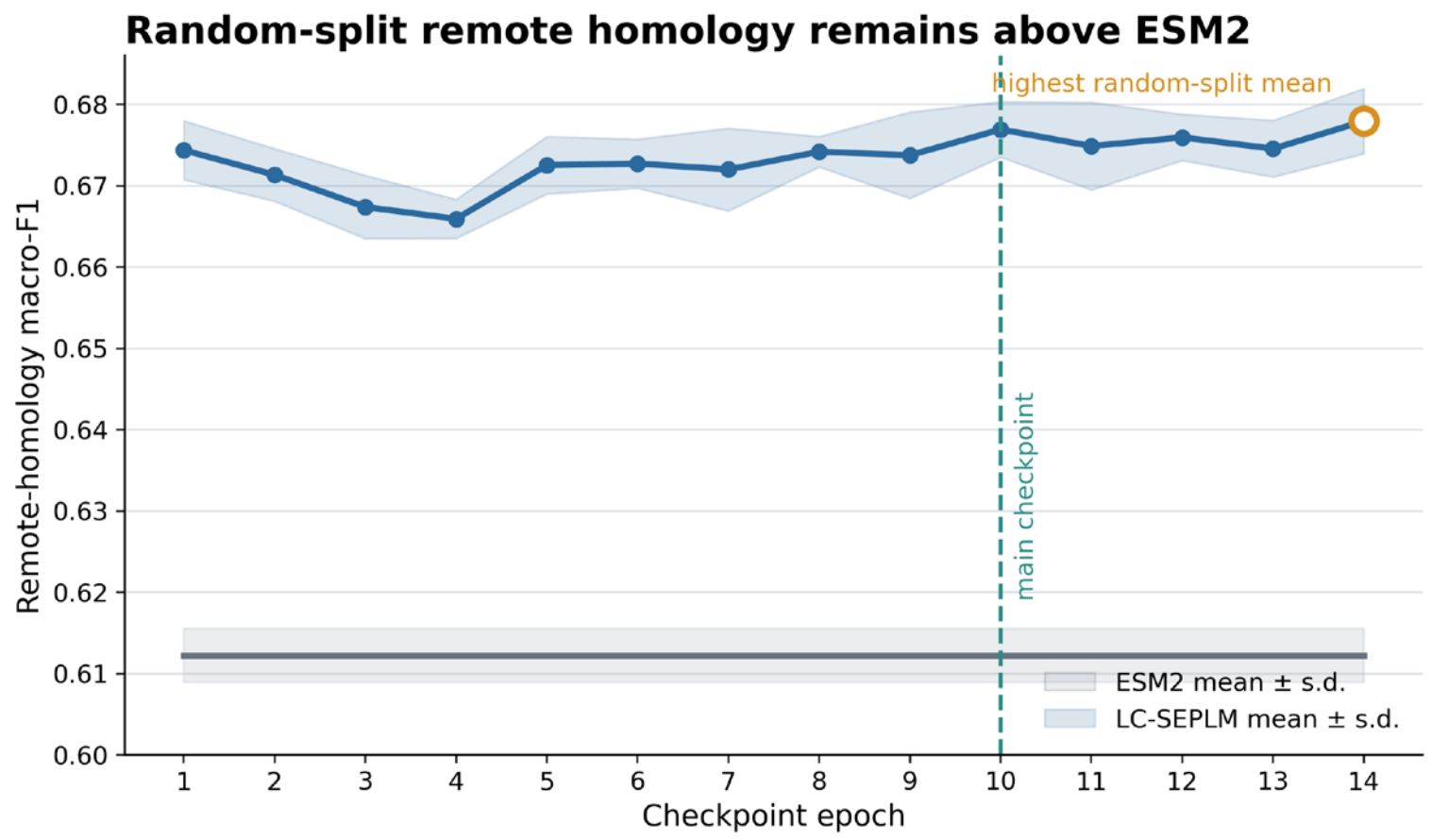

Figure 4 | Random-split remote-homology performance across checkpoints. Lines show means and shaded bands show ±1 s.d. across five classifier seeds. The vertical line marks the main epoch-10 checkpoint, and the open circle marks the highest random-split mean at epoch 14. The gray ESM2 reference is constant across checkpoints.

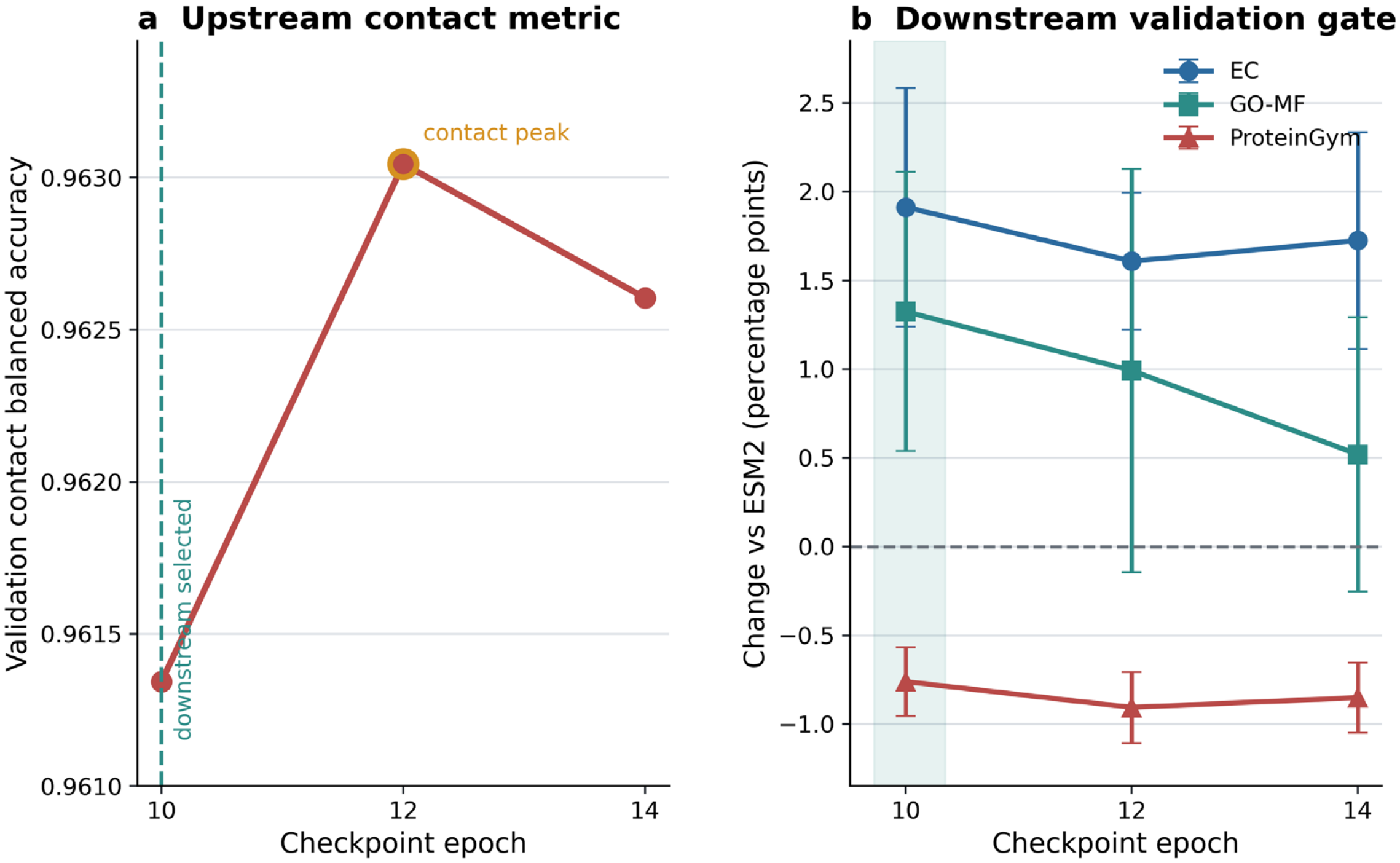

Figure 5 | Final Tail-8 checkpoint selection. a, Validation contact balanced accuracy at epochs 10, 12 and 14; the upstream metric peaks at epoch 12. b, Changes relative to ESM2 for the downstream validation gate at the same checkpoints. EC and GO-MF are means ± s.d. across three gate seeds; ProteinGym is the mean ± s.d. across 93 development assays. The shaded region marks the selected epoch 10. Gate values are used for model selection and are not the same evaluation estimates as the final comparison in Table 2.

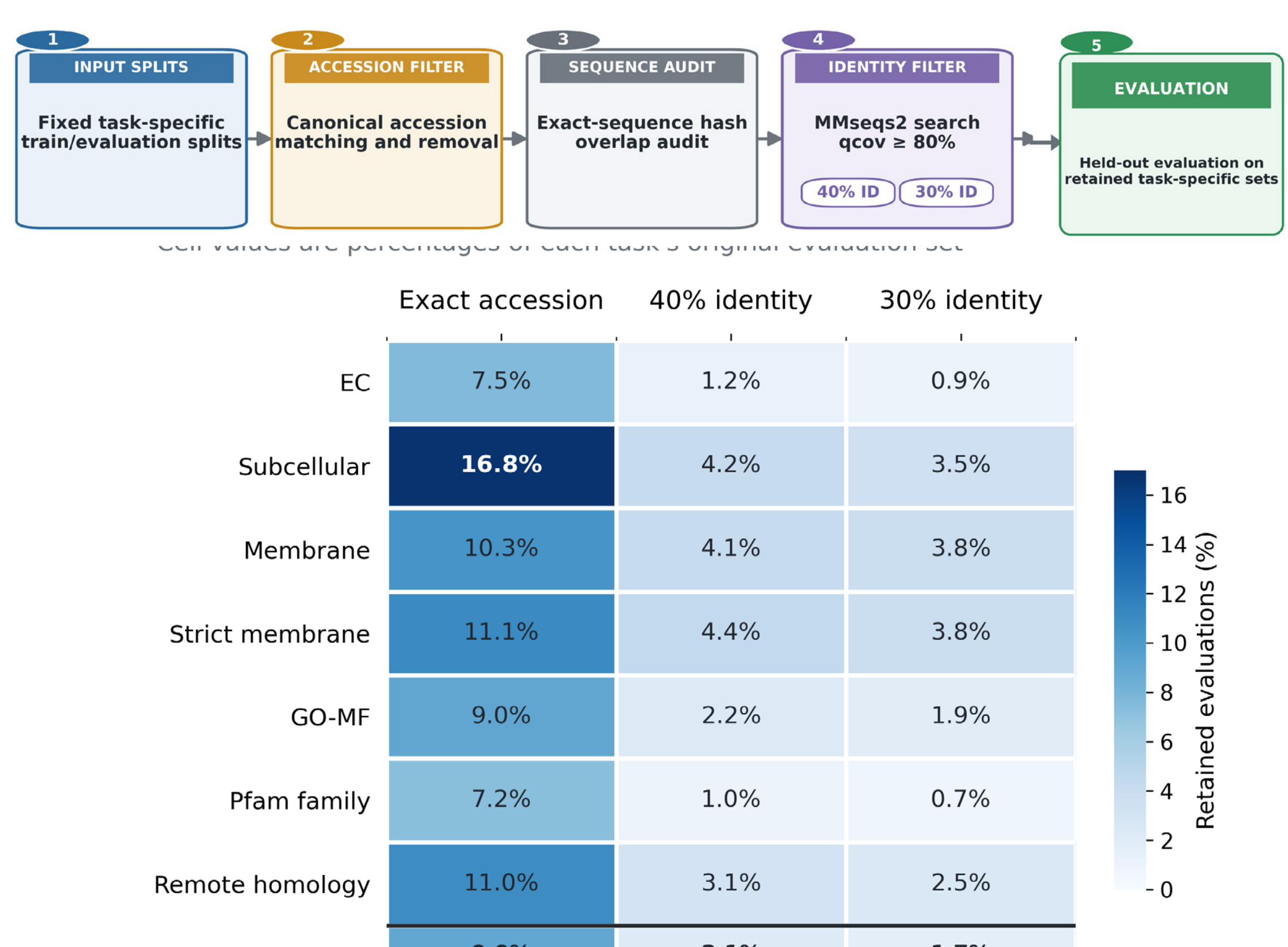

Figure 6 | Evaluation filtering for seven curated protein-level tasks. a, Structural-adaptation proteins are removed by canonical-accession matching, exact-sequence auditing and MMseqs2 identity filtering with query coverage ≥80%. b, Percentages of each original evaluation set retained after exact-accession, 40% identity and 30% identity filters. Pooled counts are summed task-level entries rather than unique proteins. HOU is not included because it has no independent validation partition.

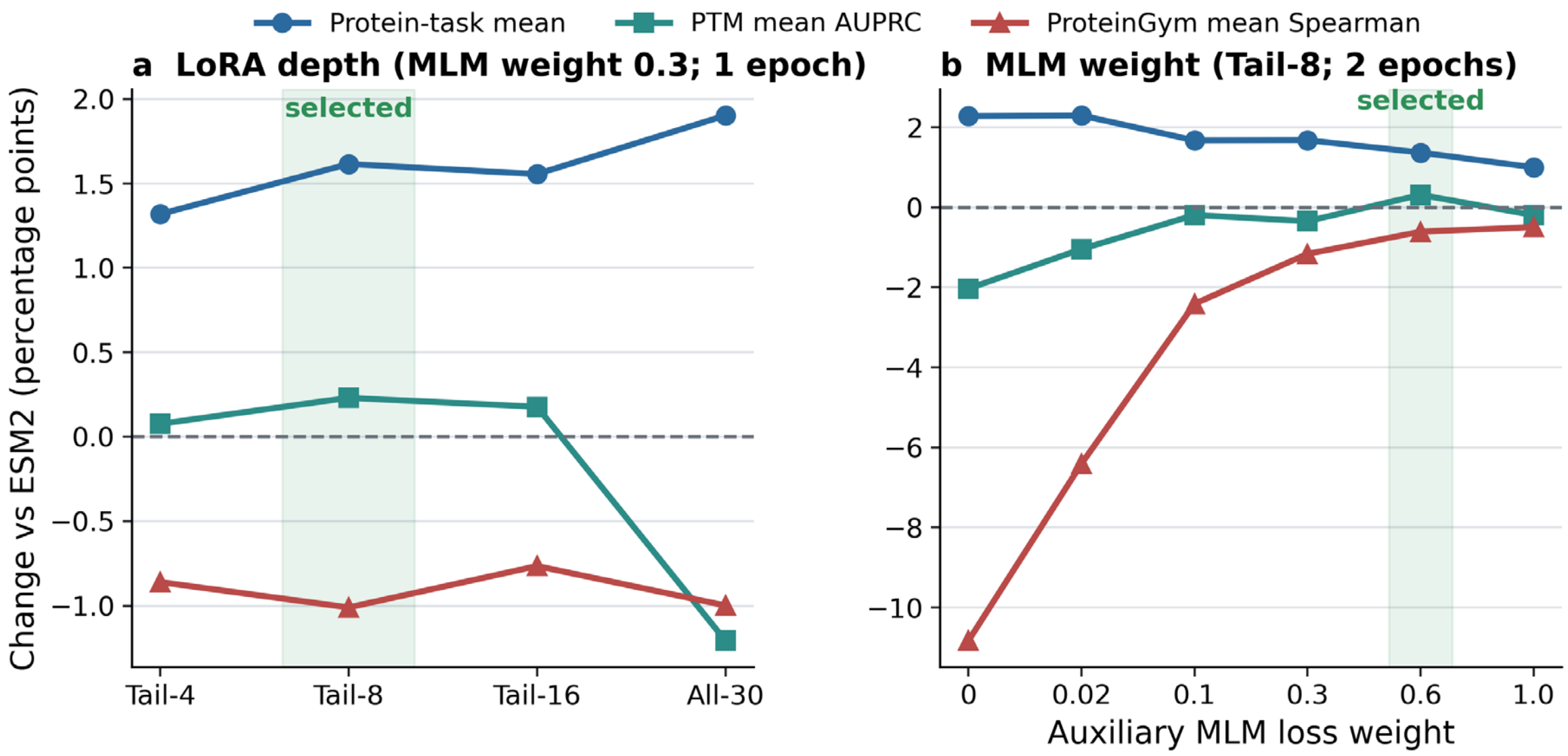

Figure 7 | LoRA-depth and auxiliary MLM-weight model-selection sweeps. a, LoRA depth was varied at MLM weight 0.3 with one epoch of training. b, MLM weight was varied with Tail-8 over two epochs. Blue shows the mean change across eight protein-level tasks, teal shows PTM mean AUPRC change and red shows ProteinGym mean Spearman change; all values are percentage-point changes relative to ESM2. Shading marks the selected Tail-8 and MLM weight 0.6 operating points. These sweeps were used for model selection rather than final confirmatory evaluation.

**Final Tail-8 transfer across eight protein-level tasks**

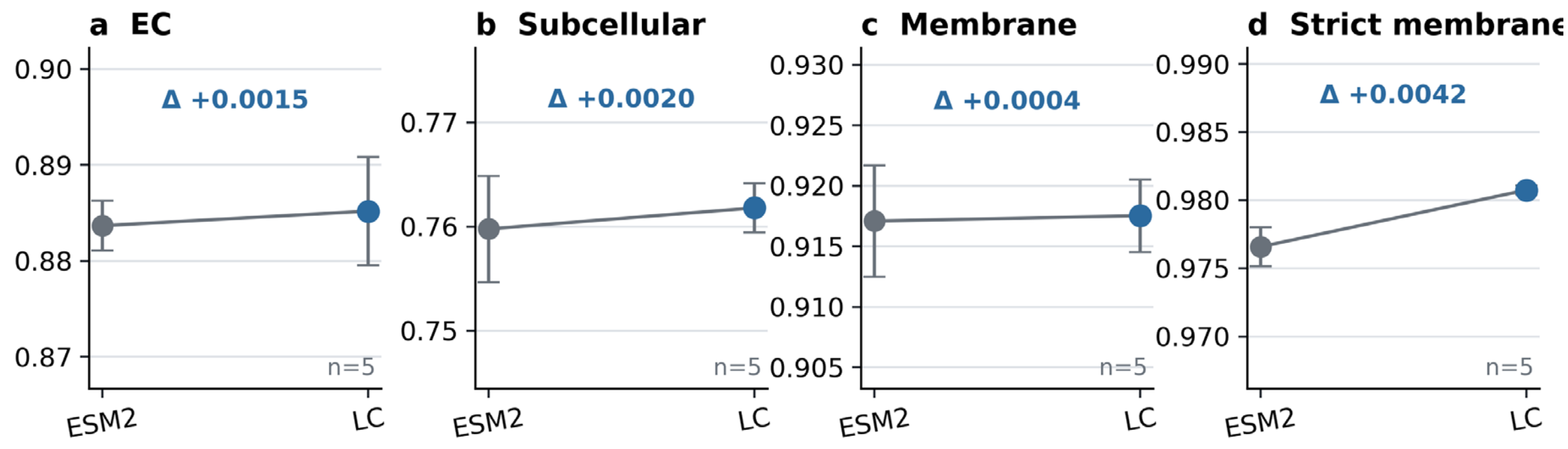

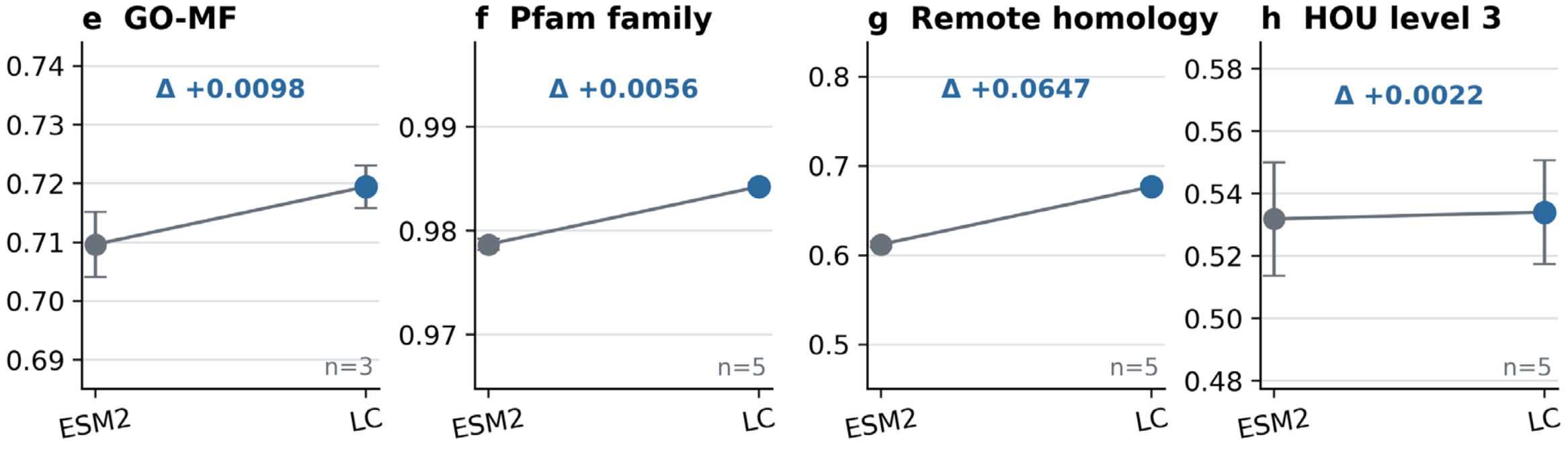

Points show mean; error bars show s.d. across downstream classifier seeds

Figure 8 | Final Tail-8 transfer across eight protein-level tasks. Each panel compares the ESM2 mean with the LC-SEPLM epoch-10 mean; error bars show s.d. across downstream classifier seeds. EC, subcellular localization, membrane classification, strict membrane classification, Pfam family, remote homology and HOU use five seeds; GO-MF uses the three-seed validation gate. The task-specific metrics are macro-F1 for EC, localization, membrane, Pfam and remote homology, and micro-F1 for GO-MF and HOU.